\documentclass[11pt, a4paper]{article} 
\usepackage[a4paper,margin=1in,]{geometry}
\usepackage{amsmath,amssymb}
\usepackage{mathtools}
\usepackage{graphicx}
\usepackage{bbm}
\usepackage{subfigure}
\usepackage{dsfont}
\usepackage{units}
\usepackage{algorithm}
\usepackage{algpseudocode}
\usepackage{wrapfig}
\usepackage{color}
\usepackage{url}
\usepackage{caption}

\usepackage{hyperref}
\hypersetup{colorlinks,
    citecolor=black,%
    filecolor=black,%
    linkcolor=black,%
    urlcolor=black}

\renewcommand{\vec}[1]{\boldsymbol{#1}} 
\newcommand{\mat}[1]{\boldsymbol{#1}} 

\newcommand{\prob}[0]{p} 
\newcommand{\E}[0]{\mathds{E}} 
\newcommand{\var}[0]{\mathrm{var}} 
\newcommand{\cov}[0]{\mathrm{cov}} 

\newcommand{\R}[0]{\mathds{R}}

\newcommand{\T}[0]{^{\top}} 
\newcommand{\inv}[0]{^{-1}} 
\newcommand{\diag}{\mathrm{diag}} 
\newcommand{\tr}{\mathrm{Tr}} 

\newcommand{\gauss}[2]{\mathcal{N}(#1,#2)}
\newcommand{\gaussBig}[2]{\mathcal{N}\left(#1,#2\right)}
\newcommand{\gaussx}[3]{\mathcal{N}(#1\,|\,#2,#3)}

\renewcommand{\d}[0]{\mathrm{d}}
\newcommand{\proj}[1]{\mathrm{proj}[#1]} 
\newcommand{\GP}[0]{\mathcal{GP}}

\newcommand{\tmu}{\tilde{\vec \mu}} %
\newcommand{\tSigma}{\tilde{\vec \Sigma}} 
\newcommand{\figspace}{\vspace{0mm}}

\newcommand{\noti}[0]{^{\setminus i}}

\renewcommand{\uparrow}{\vartriangle}
\newcommand{\back}{{\vartriangleleft}}
\newcommand{\fwd}{{\vartriangleright}}
\newcommand{\up}{{\uparrow}}
\newcommand{\nback}{{\setminus\back}}
\newcommand{\nfwd}{{\setminus\fwd}}
\newcommand{\nup}{{\setminus\up}}

\newcommand{\qfwd}{q_\vartriangleright}
\newcommand{\qback}{q_\vartriangleleft}

\newcommand{\qnfwd}{q_{\nfwd}}
\newcommand{\qnback}{q_{\nback}}
\newcommand{\qnup}{q_{\nup}}

\newcommand{\eq}{}
\newcommand{\eqs}{}

\usepackage{color}

\newcommand{\red}[1]{\textcolor{red}{#1}}
\newcommand{\blue}[1]{\textcolor{blue}{#1}}

\title{Expectation Propagation in  \\
  Gaussian Process Dynamical Systems: \\
Extended Version}

\author{
Marc Peter Deisenroth$^{*}$ \\
Department of Computing\\
Imperial College London\\
UK\\
\url{m.deisenroth@imperial.ac.uk} \\
\and
Shakir Mohamed$^*$\\
Google Deepmind\\
London\\
UK
 \\
\url{shakir@google.com} }

\date{}

\begin{document}

\maketitle

\begin{abstract}
  Rich and complex time-series data, such as those generated from
  engineering systems, financial markets, videos or neural recordings,
  are now a common feature of modern data analysis. Explaining the
  phenomena underlying these diverse data sets requires flexible and
  accurate models. In this paper, we promote Gaussian process
  dynamical systems (GPDS) as a rich model class that is appropriate
  for such analysis. In particular, we present a message passing
  algorithm for approximate inference in GPDSs based on expectation
  propagation. By posing inference as a general message passing
  problem, we iterate forward-backward smoothing.  Thus, we obtain
  more accurate posterior distributions over latent structures,
  resulting in improved predictive performance compared to
  state-of-the-art GPDS smoothers, which are special cases of our
  general message passing algorithm. Hence, we provide a unifying
  approach within which to contextualize message passing in GPDSs.
\end{abstract}

\section{Introduction}
\let\thefootnote\relax\footnotetext{*Authors contributed
  equally. Appeared in \emph{Advances in Neural
    Information Processing Systems 25, pp. 2609--2617}, 2012~\cite{Deisenroth2012d}.} 

The Kalman filter and its extensions~\cite{Anderson2005}, such as the
extended and unscented Kalman filters~\cite{Julier2004}, are
principled statistical models that have been widely used for some of
the most challenging and mission-critical applications in automatic
control, robotics, machine learning, and economics. Indeed, wherever
complex time-series are found, Kalman filters have been successfully
applied for Bayesian state estimation. However, in practice, time
series often have an unknown dynamical structure, and they are high
dimensional and noisy, violating many of the assumptions made in
established approaches for state estimation. In this paper, we look
beyond traditional linear dynamical systems and advance the state-of
the-art in state estimation by developing novel inference algorithms
for the class of nonlinear \textit{Gaussian process dynamical systems}
(GPDS).


GPDSs are non-parametric generalizations of state-space models that allow for
inference in time series, using Gaussian process (GP) probability
distributions over nonlinear transition and measurement
dynamics.  GPDSs are thus able to
capture complex dynamical structure with few assumptions, making them of broad interest. This interest
has sparked the development of general approaches for filtering and smoothing 
in GPDSs, such as ~\cite{Ko2009,Deisenroth2009a,Deisenroth2012}. In this
paper, we further develop inference algorithms for GPDSs and make the following 
contributions: (1) We develop an
iterative local message passing framework for GPDSs based on
Expectation Propagation (EP)~\cite{Minka2001, Minka2001a}, which
allows for refinement of the posterior distribution and, hence,
improved inference. (2) We show that the general message-passing
framework recovers the EP updates for existing dynamical systems as a
special case and expose the implicit modeling assumptions made in
these models. We show that EP in GPDSs encapsulates all GPDS
forward-backward smoothers~\cite{Deisenroth2012} as a special case and
transforms them into iterative algorithms yielding more accurate
inference.  



\section{Gaussian Process Dynamical Systems}
\label{sect:gpds}
Gaussian process dynamical systems are a general class of
discrete-time state-space models with
\begin{align}
  &\vec x_{t} = h(\vec x_{t-1}) + \vec
  w_t\,,\quad \vec w_t\sim\gauss{\vec 0}{\mat Q}\,, \quad
  h\sim\GP_h\,,
\label{eq:system equation}
\\
  & \vec z_t = g(\vec x_t) + \vec v_t \,,\quad \vec
  v_t\sim\gauss{\vec 0}{\mat R}\,,\quad g\sim\GP_g\,,
\label{eq:measurement equation}
\end{align}
where $t = 1,\dotsc, T$.  Here, $\vec x\in\R^D$ is a latent state that
evolves over time, and $\vec z\in\R^E$, $E\geq D$, are
measurements. We assume i.i.d. additive Gaussian system noise $\vec w$
and measurement noise $\vec v$.  The central feature of this model
class is that both the measurement function $g$ and the transition
function $h$ are not explicitly known or parametrically specified, but
instead described by probability distributions over these
functions. The function distributions are non-parametric Gaussian
processes (GPs), and we write $h\sim\GP_h$ and $g\sim\GP_g$,
respectively. 

A GP is a probability distribution $p(f)$ over functions $f$ that is
specified by a mean function $\mu_f$ and a covariance function
$k_f$~\cite{Rasmussen2006}.  Consider a set of training inputs
$\mat{X} = [\vec x_1,\dotsc,\vec x_n]\T$ and corresponding training
targets $\vec y = [y_1,\dotsc y_n]\T$, $y_i = f(\vec x_i) + w$,
$w\sim\gauss{0}{\sigma_w^2}$.  The posterior predictive distribution
at a test input $\vec x_*$ is Gaussian distributed
$\gaussx{y_*}{\mu_f(\vec x_*)}{\sigma_f^2(\vec x_*)}$ with mean
$\mu_f(\vec x_*) = \vec k_*\T\mat K\inv\vec y$ and variance
$\sigma_f^2(\vec x_*) = k_{**} - \vec k_*\T\mat K\inv\vec k_*$, where
$\vec k_*=k_f(\mat X,\vec x_*)$, $k_{**} = k_f(\vec x_*, \vec x_*)$,
and $\mat K$ is the kernel matrix. 

Since the GP is a non-parametric
model, its use in GPDSs is desirable since it results in fewer restrictive model assumptions, compared 
to dynamical systems based on parametric function approximators for the transition and
measurement functions \eqs\eqref{eq:system equation}--\eqref{eq:measurement equation}. 
  In this paper, we assume
that the GP models are trained, i.e., the training inputs and
corresponding targets as well as the GP hyperparameters are known. For
both $\GP_h$ and $\GP_g$ in the GPDS, we used zero prior mean
functions. As covariance functions $k_h$ and $k_g$ we use squared-
exponential covariance functions with automatic relevance
determination plus a noise covariance function to account for the
noise in~\eqs\eqref{eq:system equation}--\eqref{eq:measurement
  equation}.

Existing work for \emph{learning} GPDSs includes the Gaussian process
dynamical model (GPDM) \cite{Wang2008}, which tackles the challenging
task of analyzing human motion in (high-dimensional) video sequences.
More recently, variational~\cite{Damianou2011} and
EM-based~\cite{Turner2010} approaches for learning GPDS were
proposed.
%
Exact Bayesian \emph{inference}, i.e., filtering and smoothing, in
GPDSs is analytically intractable because of the dependency of the
states and measurements on previous states through the nonlinearity of
the GP.  We thus make use of approximations to infer the posterior
distributions $\prob(\mat x_t|\mat Z)$ over latent states $\mat x_t$,
$t = 1,\dotsc,T$, given a set of observations $\mat Z=\vec
z_{1:T}$. Existing approximate inference approaches for filtering and
forward-backward smoothing are based on either linearization, particle
representations, or moment matching as approximation
strategies~\cite{Ko2009,Deisenroth2009a, Deisenroth2012}.

A principled incorporation of the posterior GP model uncertainty into
inference in GPDSs is necessary, but introduces additional
uncertainty. In tracking problems where the location of an object is
not directly observed, this additional source of uncertainty can
eventually lead to losing track of the latent state. In this paper, we
address this problem and propose approximate message passing based on
EP for more accurate inference. We will show that forward-backward
smoothing in GPDSs~\cite{Deisenroth2012} benefits from the iterative
refinement scheme of EP, leading to more accurate posterior
distributions over the latent state and, hence, to more informative
predictions and improved decision making.




\section{Expectation Propagation in GPDS}
%
%
Expectation Propagation~\cite{Minka2001a, Minka2001} is a widely-used
deterministic algorithm for approximate Bayesian inference that has
been shown to be highly accurate in many problems, including sparse
regression models~\cite{Seeger2008}, GP
classification~\cite{Kuss2005}, and inference in dynamical
systems~\cite{Qi2003,Heskes2002,Toussaint2010a}.  EP is derived using
a factor-graph, in which the distribution over the latent state
$p(\vec x_t|\mat Z)$ is represented as the product of factors
$f_i(\vec x_t)$, i.e., $p(\vec x_t|\mat Z)=\prod_i f_i(\vec x_t)$.  EP
then specifies an iterative message passing algorithm in which $p(\vec
x_t|\mat Z)$ is approximated by a distribution $q (\vec x_t)=\prod_i
q_i(\vec x_t)$, using approximate messages $q_i(\vec x_t)$. In EP, $q$
and the messages $q_i$ are members of the exponential family, and $q$
is determined such that the the KL-divergence KL$(p||q)$ is minimized. EP
is provably robust for log-concave messages~\cite{Seeger2008} and
invariant under invertible variable transformations~\cite{Seeger2005}.
In practice, EP has been shown to be more accurate than competing
approximate inference methods~\cite{Kuss2005, Seeger2008}.

\begin{figure}[tb]
\centering
\subfigure{
\includegraphics[height=2cm]{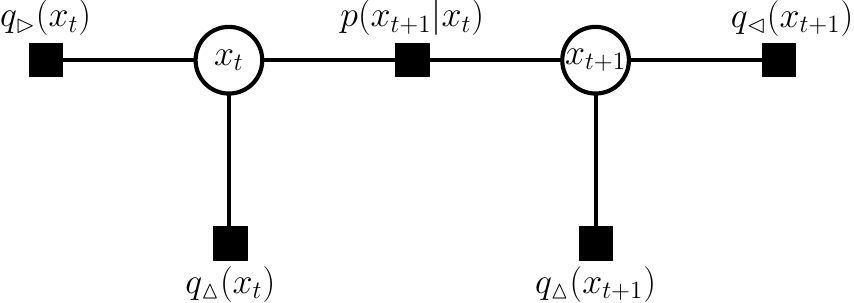}
}
\hspace{1cm}
\subfigure{
  \includegraphics[height=2cm]{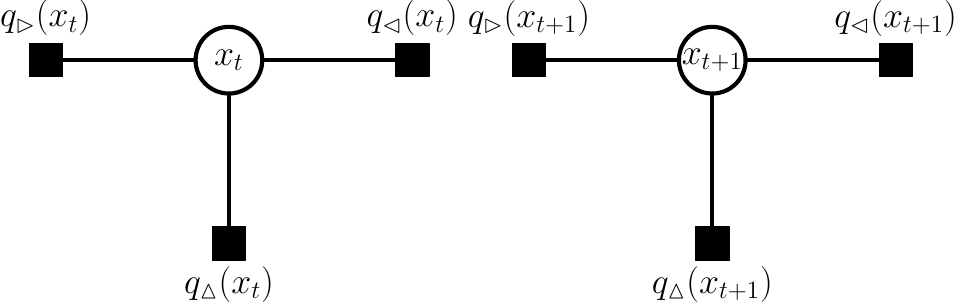}   
}
\caption{Factor graph (left) and fully factored graph (right) of a
  general dynamical system.}\label{fig:factor graph dyn system}
\figspace
\end{figure}
%
%
In the context of the dynamical system~\eq\eqref{eq:system
  equation}--\eq\eqref{eq:measurement equation}, we consider factor
graphs of the form of Fig.~\ref{fig:factor graph dyn system} with
three types of messages: forward, backward, and measurement messages,
denoted by the symbols $\fwd, \back, \up$, respectively.  For EP
inference, we assume a fully-factored graph, using which we compute the marginal
posterior distributions $p(\vec x_1|\mat Z),\dotsc, p(\vec x_T|\mat
Z)$, rather than the full joint distribution $p(\mat X|\mat Z) = p(\vec
x_1,\dotsc,\vec x_T|\mat Z)$.  Both the states $\vec x_t$ and
measurements $\vec z_t$ are continuous variables and the messages
$q_i$ are unnormalized Gaussians, i.e., $q_i(\vec x_t) = s_i
\gaussx{\vec x_t}{\vec\mu_i}{\mat\Sigma_i}$

\subsection{Implicit Linearizations Require Explicit Consideration}
\label{sect:EPGPDS}
\begin{algorithm}[t]
  \caption{Gaussian EP for Dynamical Systems}
\label{alg:ep time series}
\begin{algorithmic}[1]
  \State \textbf{Init:} Set all factors $q_i$ to $\gauss{\vec
    0}{\infty\mat I}$; Set $q(\vec x_1)=p(\vec x_1)$ and marginals
  $q(\vec x_{t\neq 1})=\gauss{\vec 0}{10^{10}\mat I}$ \Repeat \For{$t
    = 1$ to $T$} \For {all factors $q_i(\vec x_t)$, where
    $i=\fwd,\up,\back$} \State Compute cavity distribution
  $q\noti(\vec x_t)=q(\vec x_t)/q_i(\vec x_t) = \gaussx{\vec x_t}{\vec
    \mu\noti}{\mat\Sigma\noti}$ with
\begin{align}
  \mat\Sigma\noti &= (\mat\Sigma_t\inv -
  \mat\Sigma_i\inv)\inv\,,\qquad \vec \mu\noti =
  \mat\Sigma\noti(\mat\Sigma_t\inv\vec \mu_t - \mat\Sigma_i\inv\vec
  \mu_i)
\end{align} 
\State Determine moments of $f_i(\vec x_t)q\noti(\vec x_t)$, e.g., via
the derivatives of
\begin{align}
\log Z_i(\vec\mu\noti,\mat\Sigma\noti) &= \log\smallint f_i(\vec x_t) q\noti(\vec
x_t) \d\vec x_t
\label{eq:partition function}
\end{align}
\State Update the posterior $q(\vec x_t)\propto\gaussx{\vec
  x_t}{\vec\mu_t}{\mat\Sigma_t}$ and the approximate factor $q_i(\vec
x_t)$:
\begin{align}
  \vec\mu_t &= \vec\mu\noti + \mat\Sigma\noti\mat\nabla_m\T\,, \qquad
  \mat\Sigma_t = \mat\Sigma\noti - \mat\Sigma\noti(\mat \nabla_m\T\mat
  \nabla_m - 2\mat \nabla_s)\mat\Sigma\noti
  \label{eq:factored distribution update}\\
  \vec \nabla_m &\coloneqq \d\log Z_i/\d\vec\mu\noti\,,\qquad\mat
  \nabla_s \coloneqq \d\log
  Z_i/\d{\mat\Sigma\noti} \label{eq:alg_derivDef}\\
  q_i(\vec x_t) &= q(\vec x_t)/q\noti(\vec
  x_t) \label{eq:alg_margUpdate}
\end{align}
\EndFor
\EndFor
\Until Convergence or maximum number of iterations exceeded
\end{algorithmic}
\end{algorithm}
Alg.~\ref{alg:ep time series} describes the main steps of Gaussian EP
for dynamical systems.  For each node $\vec x_t$ in the fully-factored
factor graph in Fig.~\ref{fig:factor graph dyn system}, EP computes
three messages: a forward, backward, and measurement message, denoted
by $q_{\fwd}(\vec x_t)$, $q_{\back}(\vec x_t)$, and $q_{\up}(\vec
x_t)$, respectively.  The EP algorithm updates the marginal $q(\vec
x_t)$ and the messages $q_i(\vec x_t)$ in three steps. First, the
\emph{cavity distribution} $q\noti(\vec x_t)$ is computed (step 5 in
Alg.~\ref{alg:ep time series}) by removing $q_i(\vec x_t)$ from the marginal
$q(\vec x_t)$. Second, in the \emph{projection} step, the moments of
$f_i(\vec x_t)q\noti(\vec x_t)$ are computed (step 6), where $f_i$ is
the true factor. In the exponential family, the required moments can
be computed using the derivatives of the log-partition function
(normalizing constant) $\log Z_i$ of $f_i(\vec x_t)q\noti(\vec
x_t)$~\cite{Minka2001a, Minka2001, Minka2008}.  Third, the moments of
the marginal $q(\vec x_t)$ are set to the moments of $f_i(\vec
x_t)q\noti(\vec x_t)$, and the message $q_i(\vec x_t)$ is updated
(step 7). We apply this procedure repeatedly to all latent states
$\vec x_t$, $t = 1, \dotsc, T$, until convergence.

EP does not directly fit a Gaussian approximation $q_i$ to
the non-Gaussian factor $f_i$. Instead, EP determines the moments of
$q_i$ in the context of the cavity distribution such that
$q_i=\proj{f_iq\noti}/q\noti$, where $\proj{\cdot}$ is the projection
operator, returning the moments of its argument.

To update the posterior $q(\vec x_t)$ and the messages $q_i(\vec x_t)$,
EP computes the log-partition function $\log Z_i$ in
\eq\eqref{eq:partition function} to complete the projection step.
However, for nonlinear transition and measurement models
in~\eqref{eq:system equation}--\eqref{eq:measurement equation},
computing $Z_i$ involves solving integrals of the form
\begin{align}
  p(\vec a) = \int p(\vec a|\vec x_t)p(\vec x_t)\d\vec x_t =
  \int\gaussx{\vec a}{\vec m(\vec x_t)}{\mat S(\vec x_t)}\gaussx{\vec
    x_t}{\vec b}{\mat B}\d\vec x_t\,,
\label{eq:nasty integral}
\end{align}
where $\vec a=\vec z_t$ for the measurement message, or $\vec a = \vec
x_{t+1}$ for the forward and backward messages. In nonlinear dynamical
systems $\vec m(\vec x_t)$ is a nonlinear measurement or transition
function.  In GPDSs, $\vec m(\vec x_t)$ and $\mat S(\vec x_t)$ are the
corresponding predictive GP means and covariances, respectively, which
are nonlinearly related to $\vec x_t$. Because of the nonlinear
dependencies between $\vec a$ and $\vec x_t$, solving
\eq\eqref{eq:nasty integral} is analytically intractable. We propose
to approximate $p(\vec a)$ by a Gaussian distribution $\gaussx{\vec
  a}{\tmu}{\tSigma}$. This Gaussian
approximation is only correct for a linear relationship $\vec a = \mat
J\vec x_t$, where $\mat J$ is independent of $\vec x_t$. Hence, the
Gaussian approximation is an \emph{implicit linearization} of the
functional relationship between $\vec a$ and $\vec x_t$, effectively
linearizing either the transition or the measurement models.

When computing EP updates using the derivatives $\vec\nabla_m$ and $\vec\nabla_s$
according to \eq\eqref{eq:factored distribution update} it is crucial
to explicitly account for the implicit linearization assumption in the
derivatives---otherwise, the EP updates are inconsistent. For example,
in the measurement and the backward message, we directly approximate
the partition functions $Z_i$, $i\in\{\up,\back\}$ by Gaussians
$\tilde Z_i(\vec a) = \gauss{\tmu^i}{\tSigma^i}$.
%
The consistent derivatives $\d (\log \tilde Z_i)/\d\vec\mu\noti$ and
$\d (\log \tilde Z_i)/\d\mat\Sigma\noti$ of $\tilde Z_i$ with respect
to the mean and covariance of the cavity distribution $q$ are obtained
by applying the chain rule, such that
\begin{align}
  \vec\nabla_m &= \tfrac{\d\log \tilde Z_i}{\d\vec\mu\noti} =
  \tfrac{\partial\log \tilde
    Z_i}{\partial\tmu^i}\tfrac{\partial\tmu^i}{\partial\vec\mu\noti}
  = (\vec a- \tmu^i)\T(\tSigma^i)\inv \mat
  J\T\in\R^{1\times D}\,,
  \label{eq:d logZ/dm indirect}
  \\
  \vec\nabla_s &= \tfrac{\d\log\tilde Z_i}{\d\mat\Sigma\noti}
  =\tfrac{\partial\log \tilde
    Z_i}{\partial\tSigma^i}\tfrac{\partial
    \tSigma^i}{\partial\mat\Sigma\noti} =
  \tfrac{1}{2}\left(\tfrac{\partial\log \tilde
      Z_i}{\partial(\tmu^i)\T}\tfrac{\partial\log \tilde
      Z_i}{\partial\tmu^i}- (\tSigma^i)\inv\right)
  \tfrac{\partial\tSigma^i}{\partial
    \mat\Sigma\noti}\label{eq:d
    logZ/dS indirect}\in\R^{D\times D}\,,\\
  \tfrac{\partial\tmu^i}{\partial\vec\mu\noti} &= \mat
  J\T\in\R^{E\times D}\,,\quad
  \tfrac{\partial\tSigma^i}{\partial \mat\Sigma\noti} = \mat
  J\mathds{I}_4\mat J\T\in\R^{E\times E\times D\times D}\,,
\label{eq:special derivatives}
\end{align}
where $\mathds{I}_4\in\R^{D\times D\times D\times D}$ is an identity tensor.
Note that with the implicit linear model $\vec a = \mat J\vec x_t$,
the derivatives $\partial\tmu^i/\partial\mat\Sigma\noti$ and
$\partial\tSigma^i/\partial\vec\mu\noti$ vanish. 
%
Although we approximate $Z_i$ by a Gaussian $\tilde Z_i$, we are still
free to choose a method of computing its mean $\tmu^i$ and
covariance matrix $\tSigma^i$, which also influences the
computation of $\mat
J=\partial(\tmu^i)/\partial\vec\mu\noti$. However, even if
$\tmu^i$ and $\tSigma^i$ are general functions of
$\vec\mu\noti$ and $\mat\Sigma\noti$, the derivatives
$\partial\tmu^i/\partial\vec\mu\noti$ and
$\partial\tSigma^i/\partial\vec\Sigma\noti$ must equal the
corresponding partial derivatives in \eq\eqref{eq:special
  derivatives}, and $\partial\tmu^i/\partial\mat\Sigma\noti$
and $\partial\tSigma^i/\partial\vec\mu\noti$ must be set to
$\vec 0$. Hence, the implicit linearization expressed by the Gaussian
approximation $\tilde Z_i$ must be explicitly taken into account in
the derivatives to guarantee consistent EP updates.


\subsection{Messages in Gaussian Process Dynamical Systems}
\label{sec:messages}
We now describe each the messages needed for inference in GPDSs, and outline the
 approximations required to compute the partition function in
\eq\eqref{eq:partition function}. Updating a message requires a
\textit{projection} to compute the moments of the new posterior
marginal $q(\vec x_t)$, followed by a Gaussian division to update
the message itself. For the projection step, we compute approximate
partition functions $\tilde Z_i$, where
\mbox{$i\in\{\up,\fwd,\back\}$}. Using the derivatives $\d\log\tilde
Z_i/\d\vec\mu_t\noti$ and $\d\log\tilde Z_i/\d\mat\Sigma_t\noti$, we
update the marginal $q(\vec x_t)$, see \eq\eqref{eq:factored
  distribution update}.

\paragraph{Measurement Message}
For the measurement message in a GPDS, the partition function is
\begin{align}
  Z_\uparrow(\vec\mu_t^\nup,\mat \Sigma_t^\nup) & = \int
  f_\uparrow(\vec x_t)\qnup(\vec x_t)\d\vec x_t \propto \int
  f_\uparrow(\vec x_t) \gaussx{\vec
    x_t}{\vec\mu_t^\nup}{\mat\Sigma_t^\nup}\d\vec x_t\,,
  \label{eq:Z-measurement} \\
  f_\uparrow(\vec x_t) &= \prob(\vec z_t|\vec x_t) = \gaussx{\vec
    z_t}{\vec\mu_g(\vec x_t)}{\mat\Sigma_g(\vec x_t)},
   \label{eq:true factor measurement message}
\end{align}
where $f_\uparrow$ is the true measurement factor, and $\vec\mu_g(\vec
x_t)$ and $\mat\Sigma_g(\vec x_t)$ are the predictive mean and
covariance of the measurement GP $\GP_g$.  In
\eq\eqref{eq:Z-measurement}, we made it explicit that $Z_\up$ depends
on the moments $\vec\mu_t^\nup$ and $\mat\Sigma_t^\nup$ of the cavity
distribution $\qnup(\vec x_t)$.  The integral in
\eq\eqref{eq:Z-measurement} is of the form \eq\eqref{eq:nasty
  integral}, but is intractable since solving it corresponds to a GP
prediction with uncertain inputs \cite{Rasmussen2006} which is no longer Gaussian.
However, the mean and covariance of a Gaussian approximation $\tilde
Z_\up$ to $Z_\up$ can be computed analytically: either using exact moment
matching~\cite{Quinonero-Candela2003a,Deisenroth2009a}, or
approximately by expected linearization of the posterior
GP~\cite{Ko2009}; details are given in the Appendix. The
moments of $\tilde Z_\up$ are also functions of the mean $\vec\mu_t^\nup$
and variance $\mat\Sigma_t^\nup$ of the cavity distribution.
By taking the linearization assumption of the Gaussian approximation
into account explicitly (here, we implicitly linearize $\GP_g$) when
computing the derivatives, the EP updates remain consistent, see
Sec.~\ref{sect:EPGPDS}.

\paragraph{Backward Message}
To update the backward message $\qback(\vec x_t)$,
we require the partition function
\begin{align}
  &\hspace{-2mm}Z_\back(\vec\mu_t^\nback,\mat \Sigma_t^\nback)  = \int
  f_\back(\vec x_t)\qnback(\vec x_t)\d\vec x_t \propto \int
  f_\back(\vec x_t) \gaussx{\vec
    x_t}{\vec\mu_t^\nback}{\mat\Sigma_t^\nback}\d\vec x_t\,,
  \label{eq:backward message close approximation} \\
  &\hspace{-2mm}f_\back(\vec x_t)\!=\!  \int\prob(\vec x_{t+1}|\vec
  x_t)\qnfwd(\vec x_{t+1}) \d\vec x_{t+1}\! =\!  \int\gaussx{\vec
    x_{t+1}}{\vec \mu_h(\vec x_t)}{\mat\Sigma_h(\vec x_t)}\qnfwd(\vec
  x_{t+1}) \d\vec x_{t+1}\,.
\label{eq:trueFactorBackwardMessage}
\end{align}
Here, the true factor $f_\back(\vec x_t)$ in
\eq\eqref{eq:trueFactorBackwardMessage} takes into account the
coupling between $\vec x_t$ and $\vec x_{t+1}$, which was lost in
assuming the full factorization in Fig.~\ref{fig:factor graph dyn
  system}. The predictive mean and covariance of $\GP_h$ are denoted
 $\vec\mu_h(\vec x_t)$ and $\mat\Sigma_h(\vec x_t)$, respectively.
Using \eq\eqref{eq:trueFactorBackwardMessage} in \eq\eqref{eq:backward
  message close approximation} and reordering the integration yields
\begin{align}
  Z_\back (\vec\mu_t^\nback,\mat \Sigma_t^\nback)&\propto \int q_\nfwd(\vec x_{t+1})
  \int\prob(\vec x_{t+1}|\vec x_t)\qnback(\vec x_t) \d\vec x_{t} \d\vec
  x_{t+1}\,.
\label{eq:Z backward message}
\end{align}
We approximate the inner integral in \eq\eqref{eq:Z backward message},
which is of the form \eq\eqref{eq:nasty integral}, by $\gaussx{\vec
  x_{t+1}}{\tmu^\nback}{\tSigma^\nback}$ by moment
matching~\cite{Quinonero-Candela2003a}, for instance. Note that
$\tmu^\nback$ and $\tSigma^\nback$ are functions
of $\vec\mu_t^\nback$ and $\mat\Sigma_t^\nback$. This Gaussian
approximation implicitly linearizes $\GP_h$. Now, \eq\eqref{eq:Z
  backward message} can be computed analytically, and we obtain a
Gaussian approximation \mbox{$\tilde Z_\back =
  \gaussx{\vec\mu_{t+1}^\nfwd}{\tmu^\nback}{\tSigma^\nback +
    \mat\Sigma_{t+1}^\nfwd}$} of $Z_\back$ that allows us to update
the moments of $q(\vec x_t)$ and the message $\qback (\vec x_t)$.


\paragraph{Forward Message}
Similarly, for the forward message, the projection step involves
computing the partition function
\begin{align}
   &Z_\fwd(\vec\mu_{t}^\nfwd,\mat \Sigma_{t}^\nfwd)= \int
  f_\fwd(\vec x_{t})\qnfwd(\vec x_{t})\d\vec x_{t}=\int
  f_\fwd(\vec x_{t}) \gaussx{\vec
    x_{t}}{\vec\mu_{t}^\nfwd}{\mat\Sigma_{t}^\nfwd}\d\vec x_{t},
  \label{eq:forward message close approximation} \\
  & f_\fwd(\vec x_{t})  = \int\prob(\vec x_{t}|\vec
  x_{t-1})\qnback(\vec x_{t-1}) 
  \d\vec x_{t-1} = \int\gaussx{\vec x_{t}}{\vec\mu_h(\vec
    x_{t-1})}{\mat\Sigma_h(\vec x_{t-1})}\qnback(\vec x_{t-1}) \d\vec x_{t-1}\,,
\label{eq:trueFactorForwardMessage}\nonumber
\end{align}
where the true factor $f_\fwd(\vec x_{t})$ takes into account the
coupling between $\vec x_{t-1}$ and $\vec x_t$, see
Fig.~\ref{fig:factor graph dyn system}.
%
Here, the true factor $f_\fwd(\vec x_{t})$ is of the
form~\eq\eqref{eq:nasty integral}. We propose to approximate
$f_\fwd(\vec x_t)$ directly by a Gaussian $q_\fwd(\vec
x_{t})\propto\gauss{\tmu^\fwd}{\tSigma^\fwd}$. This
approximation implicitly linearizes $\GP_h$.  We obtain the updated
posterior $q(\vec x_{t})$ by Gaussian multiplication, i.e., $q(\vec
x_{t}) \propto \qfwd(\vec x_{t})\qnfwd(\vec x_{t})$.  With this
approximation we do not update the forward message in context, i.e.,
the true factor $f_\fwd(\vec x_t)$ is directly approximated instead of
the product $f_\fwd(\vec x_t) \qnfwd(\vec x_t)$, which can result in
suboptimal approximation.

\subsection{EP Updates for General Gaussian Smoothers}

We can interpret the EP computations in the context of
classical Gaussian filtering and smoothing~\cite{Anderson2005}.
During the \emph{forward sweep}, the marginal $q(\vec
x_t)=\qnback(\vec x_t)$ corresponds to the filter distribution $p(\vec
x_t|\vec z_{1:t})$. Moreover, the cavity distribution $\qnup (\vec
x_t)$ corresponds to the time update $p(\vec x_t|\vec z_{1:t-1})$. In
the \emph{backward sweep}, the marginal $q(\vec x_t)$ is the smoothing
distribution $p(\vec x_t|\mat Z)$, incorporating the measurements of
the entire time series. The mean and covariance of $\tilde Z_\back$
can be interpreted as the mean and covariance of the time update
$\prob(\vec x_{t+1}|\vec z_{1:t})$.

Updating the moments of the posterior $q(\vec x_t)$ via the
derivatives of the log-partition function recovers exactly the
standard Gaussian EP updates in dynamical systems described by Qi and Minka~\cite{Qi2003}.  For
example, when incorporating an updated \emph{measurement message}, the
moments in \eq\eqref{eq:factored distribution update} can also be
written as $\vec\mu_t = \vec\mu_t^\nup + \mat K(\vec z_t -
\vec\mu_z^\nup)$ and $\mat\Sigma_t = \mat\Sigma_t^\nup - \mat
K\mat\Sigma_t^{zx\nup} $, respectively, where $\mat\Sigma_t^{xz\nup} =
\cov[\vec x_t^\nup, \vec z_t^\nup]$ and $\mat K =
\mat\Sigma_t^{xz\nup}(\mat\Sigma_z^\nup)\inv$. Here,
$\vec\mu_z^\nup=\E[g(\vec x_t)]$ and $\mat\Sigma_z^\nup=\cov[g(\vec
x_t)]+\mat R$, where $\vec x_t\sim \qnup(\vec x_t)$.  Similarly, the updated
moments of $q(\vec x_t)$ with a new \emph{backward message} via
\eq\eqref{eq:factored distribution update} correspond to the
updates~\cite{Qi2003} $ \vec\mu_{t} = \vec \mu_t^\nback + \mat
L(\vec\mu_{t+1} - \vec\mu_{t+1}^\nback)$ and $ \mat\Sigma_{t} =
\mat\Sigma_t^\nback + \mat L(\mat\Sigma_{t+1} -
\mat\Sigma_{t+1}^\nback)\mat L\T$, where $\mat L = \cov[\vec
x_t^\nback, \vec x_{t+1}^\nback](\mat\Sigma_{t+1}^\nback)\inv$. Here,
we defined $\vec\mu_{t+1}^\nback=\E[h(\vec x_t)]$ and
$\mat\Sigma_{t+1}^\nback=\cov[h(\vec x_t)]+\mat Q$, where $\vec
x_t\sim\qnback(\vec x_t)$.

The iterative message-passing algorithm in Alg.~\ref{alg:ep time
  series} provides an EP-based generalization and a unifying view of
existing approaches for smoothing in dynamical systems, e.g.,
(Extended\slash Unscented\slash Cubature) Kalman smoothing and the
corresponding GPDS smoothers~\cite{Deisenroth2012}.  Computing the
messages via the derivatives of the approximate log-partition
functions $\log\tilde Z_i$ recovers not only standard EP updates in
dynamical systems~\cite{Qi2003}, but also the standard Kalman
smoothing updates~\cite{Anderson2005}.

Using any prediction method (e.g., unscented transformation,
linearization), we can compute Gaussian approximations of
\eq\eqref{eq:nasty integral}. This influences the computation of
$\log\tilde Z_i$ and its derivatives with respect to the moments of
the cavity distribution, see \eqs\eqref{eq:d logZ/dm
  indirect}--\eqref{eq:d logZ/dS indirect}. Hence, our message-passing
formulation is also general as it includes all conceivable Gaussian
filters\slash smoothers in (GP)DSs, solely depending on the prediction
technique used.

%




\section{Experimental Results}
We evaluated our proposed EP-based message passing algorithm on three
data sets: a synthetic data set, a low-dimensional simulated
mechanical system with control inputs, and a high-dimensional
motion-capture data set.  We compared to existing state-of-the-art
forward-backward smoothers in GPDSs, specifically the
GPEKS~\cite{Ko2009}, which is based on the expected linearization of
the GP models, and the GPADS~\cite{Deisenroth2012}, which uses
moment-matching. We refer to our EP generalizations of these methods
as EP-GPEKS and EP-GPADS. 

%
In all our experiments, we evaluated the inference methods using test
sequences of measurements $\mat Z=[\vec z_1,\dotsc,\vec z_T]$.  We
report the negative log-likelihood of predicted measurements using the
observed test sequence (NLL$_z$).  Whenever available, we also
compared the inferred posterior distribution $q(\mat X)\approx p(\mat
X|\mat Z)$ of the latent states with the underlying ground truth using
the average negative log-likelihood (NLL$_x$) and Mean Absolute Errors
(MAE$_x$).  We terminated EP after 100 iterations or when the average
norms of the differences of the means and covariances of $q(\mat X)$
in two subsequent EP iterations were smaller than $10^{-6}$.


\subsection{Synthetic Data}
We considered the nonlinear dynamical system
\begin{align*}
x_{t+1} = 4\sin(x_t) + w \,,\quad  w\sim\gauss{0}{0.1^2} \,,\qquad
 z_t = 4\sin(x_t) + v \,,\quad v\sim\gauss{0}{0.1^2}\,.
\end{align*}
We used $p(x_1) = \gauss{0}{1}$ as a prior on the initial latent
state. We assumed access to the latent state and trained the dynamics and
measurement GPs using 30 randomly generated points, resulting in a
model with a substantial amount of posterior model uncertainty. The
length of the test trajectory used was $T = 20$ time steps. 

\begin{figure}[tb]
  \centering \subfigure[Example trajectory distributions with
  95\% confidence bounds.]{
\includegraphics[width = 0.48\hsize]{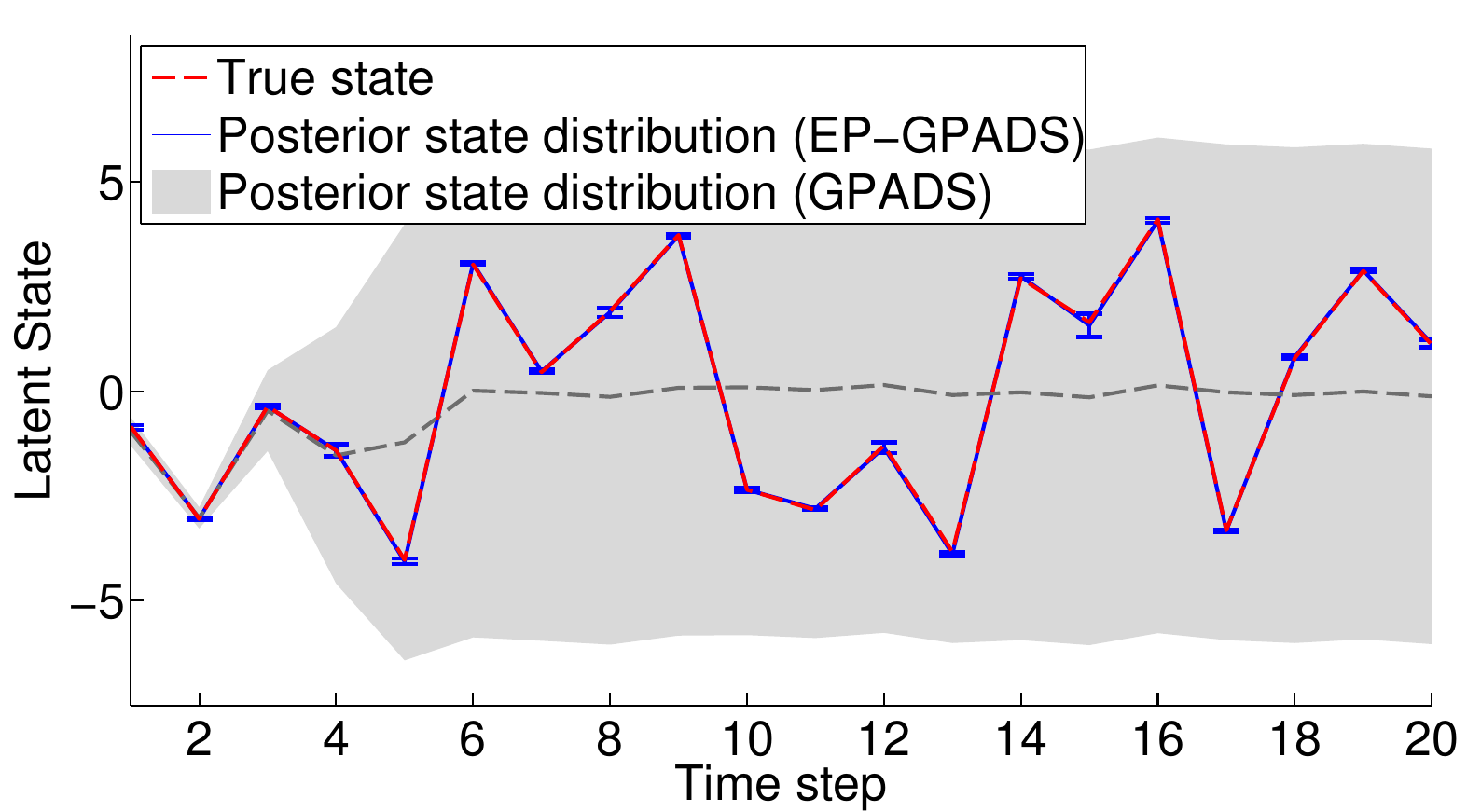}
\label{fig:1d example ADF}
}
\hfill
 \subfigure[Average NLL$_x$ as a function of the EP iteration with
twice the standard error.]{
\includegraphics[width = 0.48\hsize]{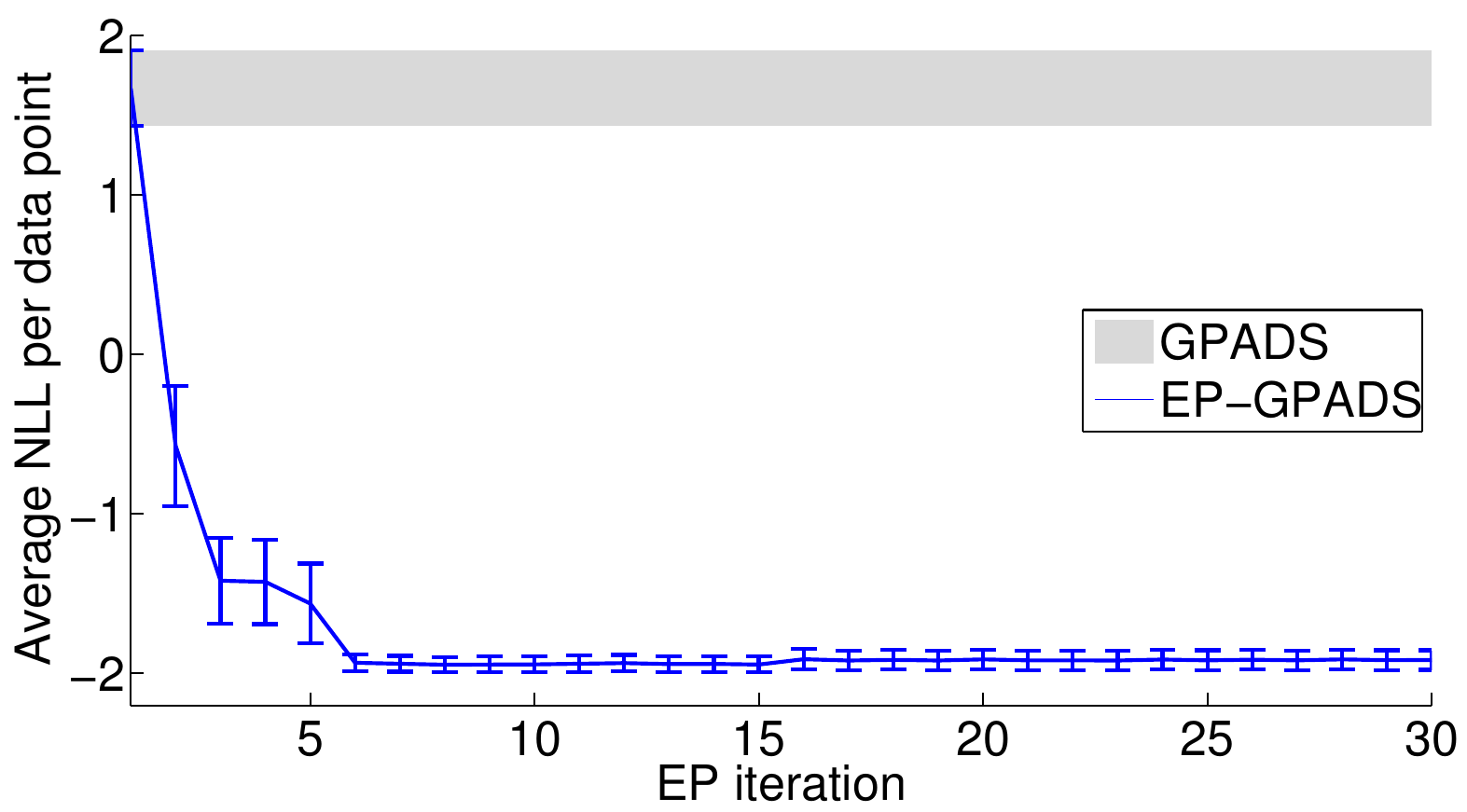}
\label{fig:1d NLL ADF}
}
\caption{\subref{fig:1d example ADF} Posterior latent state distributions
  using EP-GPADS (blue) and the GPADS (gray). The ground truth is shown in
  red (dashed).  The GPADS quickly loses track of the period of the
  state revealed by the large posterior uncertainty. EP with moment
  matching (EP-GPADS) in the GPDS 
  iteratively refines the GPADS posterior and can closely follow the true
  latent state trajectory.  \subref{fig:1d NLL ADF} Average NLL$_x$
  per data point in latent space with standard errors of the posterior
  state distributions computed by the GPADS and the EP-GPADS as a
  function of EP iterations.
\figspace
}
\end{figure}

\begin{table}[tb]
\centering
\captionof{table}{Performance comparison on the synthetic data set. Lower
  values are better.}
\label{tab:NLL, MAE and LPU in latent space 1D example}
\scalebox{0.75}{
\begin{tabular}{c|c c |c c |c c}
  & EKS & EP-EKS & {GPEKS}   & {EP-GPEKS} & GPADS & {EP-GPADS} \\
  \hline
  NLL$_x$ & $-2.04\pm 0.07$ &  $-2.17\pm 0.04$ &  $-1.67\pm 0.22$ &
  $-1.87\pm 0.14$ &
  $\red{\bf+1.67\pm  0.37}$    & $\blue{\bf -1.91\pm 0.10}$\\
  MAE$_x$ &  $0.03\pm 2.0\times 10^{-3}$ &   $0.03\pm 2.0\times 10^{-3}$ &
  $0.04\pm 4.6\times 10^{-2} $ 
  &  $0.04\pm 4.6\times 10^{-2} $
  &   $\red{\bf 1.79\pm 0.21}$    
  &  $\blue{\bf 0.04\pm 4\times 10^{-3}}$\\
  NLL$_z$
  & $ -0.69 \pm 0.11$& $ -0.73 \pm 0.11$& $ -0.75 \pm 0.08$& $ -0.81 \pm 0.07$& $ 1.93 \pm 0.28$& $ -0.77 \pm 0.07$ 
\end{tabular}
}
\end{table}
%
Tab.~\ref{tab:NLL, MAE and LPU in latent space 1D example} reports the
quality of the inferred posterior distributions of the latent state
trajectories using the average NLL$_x$, MAE$_x$, and NLL$_z$ (with
standard errors), averaged over 10 independent scenarios.  
For this  dataset, we also compared to the Extended Kalman Smoother (EKS) and an EP-iterated EKS
(EP-EKS), as models which make use of the known dynamics.
Iterated forward-backward smoothing with EP (EP-EKS, EP-GPEKS, EP-GPADS)
improved the smoothing posteriors using a single sweep only (EKS,
GPEKS, GPADS). 
The GPADS had poor performance across all our evaluation criteria for
two reasons: First, the GPs were trained using few data points,
resulting in posterior distributions with a high degree of
uncertainty. Second, predictive variances using moment-matching are
generally conservative and increased the uncertainty even
further.  This uncertainty caused the GPADS to quickly lose track of
the period of the state, as shown in Fig.~\ref{fig:1d example ADF}. By
iterating forward-backward smoothing using EP (EP-GPADS), the
posteriors $p(\vec x_t|\mat Z)$ were iteratively refined, and the
latent state could be followed closely as indicated by both the small
blue error bars in Fig.~\ref{fig:1d example ADF} and all performance
measures in Tab.~\ref{tab:NLL, MAE and LPU in latent space 1D
  example}.
%
EP smoothing typically required a small number of iterations for the
inferred posterior distribution to closely track the true state,
Fig.~\ref{fig:1d NLL ADF}. On average, EP required fewer than 10
iterations to converge to a good solution in which the mean of the
latent-state posterior closely matched the ground truth.
%
\subsection{Pendulum Tracking}
%
We considered a pendulum tracking problem to demonstrate GPDS inference
in multidimensional settings, as well as the ability to handle control inputs.  
The state $\vec x$
of the system is given by the angle $\phi$ measured from being
upright and the angular velocity $\dot\phi$. The pendulum used has a mass
of $\unit[1]{kg}$ and a length of $\unit[1]{m}$, and random torques
$u\in [-2,2]\,\unit{Nm}$ were applied for a duration $\unit[200]{ms}$
(zero-order-hold control). 
The system noise covariance was set to
$\mat\Sigma_w = \diag(0.3^2, 0.1^2).$ The state was measured
indirectly by two bearings sensors with coordinates $(x_1,
y_1)=(-2,0)$ and $(x_2, y_2)=(-0.5,-0.5)$, respectively, according to
$ \vec z = [z_1, z_2]\T+\vec v\,,\,\, \vec v\sim\gaussBig{\vec
  0}{\diag(0.1^2, 0.05^2)}$ with $ z_i =\arctan \big( \tfrac{\sin\phi
  - y_i}{\cos\phi - x_i} \big)$, $i = 1,2$.
We trained the GP models using 4 randomly generated trajectories of
length $T=20$ time steps, starting from an initial state distribution
$\prob(\vec x_1) = \gauss{\vec 0}{\diag(\pi^2/16^2, 0.5^2)}$ around
the upright position. For testing, we generated 12 random trajectories
starting from $\prob(\vec x_1)$. 
%
%
%
%
%
%

\begin{wrapfigure}{r}{0.6\textwidth}
\vspace{-4mm}
\centering
\captionof{table}{Performance comparison on the pendulum-swing
  data. Lower values are better.}
\label{tab:comparison_mocap_pendulum}
\scalebox{0.9}{
\begin{tabular}{l|ccc }
 & NLL$_x$ & MAE$_x$ & NLL$_z$\\
\hline
GPEKS  &
$\red{\bf -0.35 \pm 0.39}$ & 
$0.30 \pm 0.02 $ & 
$-2.41 \pm 0.047 $\\
EP-GPEKS  &
$\red{\bf -0.33\pm 0.44}$ &  
$0.31\pm 0.02 $ & 
$-2.39\pm 0.038$ \\
GPADS   & 
$\blue{\bf -0.80 \pm 0.06 }$ & 
$ 0.30 \pm 0.02 $ & 
$-2.37 \pm 0.042 $\\
EP-GPADS  & 
$\blue{\bf -0.85\pm  0.05}$ & 
$0.29\pm 0.02$ &  
$-2.40 \pm 0.037 $
\end{tabular}
}
\vspace{-2mm}
\end{wrapfigure}
Tab.~\ref{tab:comparison_mocap_pendulum} summarizes the performance of
the various inference methods.  Generally, the (EP-)GPADS performed
better than the (EP-)GPEKS across all performance measures. This
indicates that the (EP-)GPEKS suffered from overconfident posteriors
compared to (EP-)GPADS, which is especially pronounced in the
degrading NLL$_x$ values with increasing EP iterations and the
relatively high standard errors.
%
In about 20\% of the test cases, the inference methods based on explicit
linearization of the posterior mean function (GPEKS and EP-GPEKS) ran
into numerical problems typical of
linearizations~\cite{Deisenroth2012}, i.e., overconfident posterior
distributions that caused numerical problems. We excluded these runs
from the results in Tab.~\ref{tab:comparison_mocap_pendulum}. The
inference algorithms based on moment matching (GPADS and EP-GPADS)
were numerically stable as their predictions are typically more
coherent due to conservative approximations of moment matching.

\subsection{Motion Capture Data}
%
We considered motion capture data (from
\url{http://mocap.cs.cmu.edu/}, subject 64) containing 10 trials of
golf swings recorded at $\unit[120]{Hz}$, which we subsampled to
$\unit[20]{Hz}$. After removing observation dimensions with no
variability we were left with observations $\vec z_t\in\R^{56}$, which
were then whitened as a pre-processing step. For trials 1--7 (403 data
points), we used the GPDM~\cite{Wang2008} to learn MAP estimates of
the latent states $\vec x_t\in\R^3$. These estimated latent states and
their corresponding observations are used to train the GP models
$\GP_f$ and $\GP_g$.  Trials 8--10 were used as test data without
ground truth labels. The GPDM~\cite{Wang2008} focuses on learning a
GPDS; we are interested in good approximate inference in these models.

\begin{figure}[tb]
\centering
\includegraphics[width = 0.65\hsize]{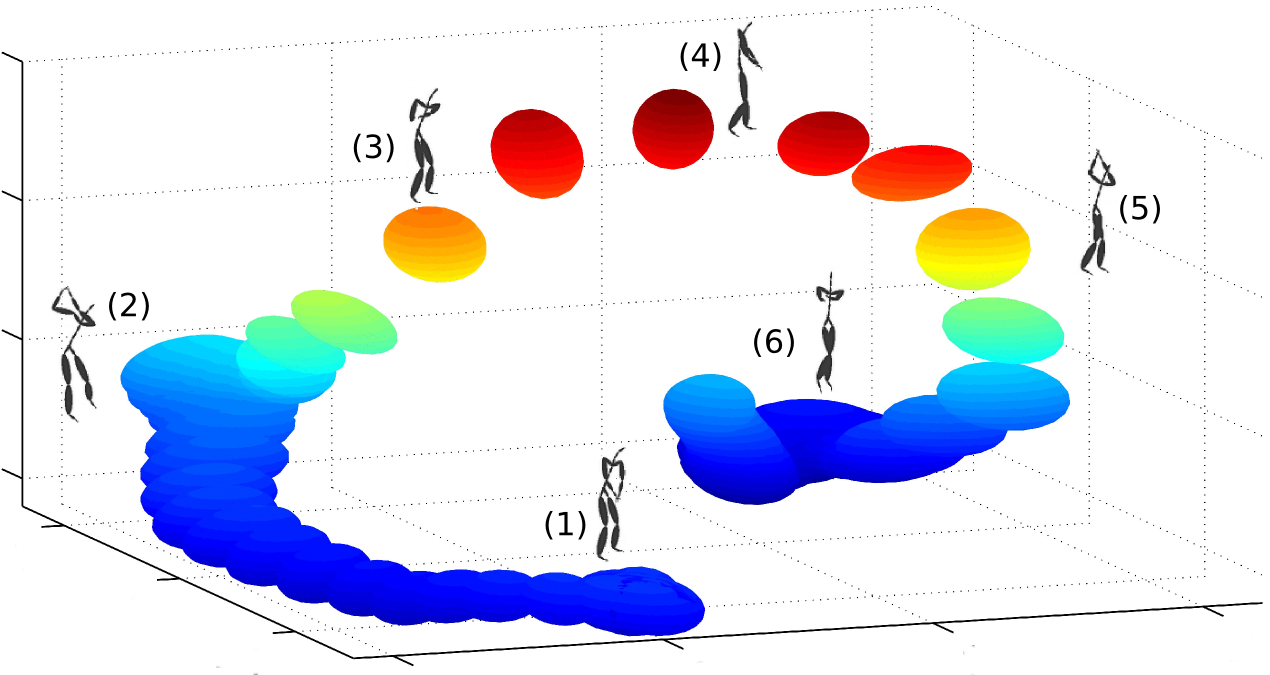}
\caption{Latent space posterior distribution (95\% confidence
  ellipsoids) of a test trajectory of the golf-swing motion capture
  data. The further the ellipsoids are separated the faster the
  movement. }
\label{fig:latent space posterior golf}
\figspace
\end{figure}
%
%
Fig.~\ref{fig:latent space posterior golf} shows the latent-state
posterior distribution of a single test sequence (trial 10) obtained
from the EP-GPADS.  The most significant
prediction errors in observed space occurred in the region
corresponding to the yellow\slash red ellipsoids, which is a
low-dimensional embedding of the motion when the golf player hits the
ball, i.e., the periods of high acceleration (poses 3--5).

Tab.~\ref{fig:mocap results} summarizes the results of
inference on the golf data set in all test trials: Iterating
forward-backward smoothing by means of EP improved the inferred
posterior distributions over the latent states.  The posterior
distributions in latent space inferred by the EP-GPEKS were tighter
than the ones inferred by the EP-GPADS. The NLL$_z$-values suffered a
bit from this overconfidence, but the predictive performance of the
EP-GPADS and EP-GPEKS were similar. Generally, inference was more
difficult in areas with fast movements (poses 3--5 in
Fig.~\ref{fig:latent space posterior golf}) where training data were
sparse.



\begin{wrapfigure}{r}{0.6\textwidth}
\vspace{-4mm}
\centering
  \captionof{table}{Average inference performance (NLL$_z$, motion capture data
    set). Lower values are better.}
\label{fig:mocap results}
\scalebox{0.9}{
\begin{tabular}{l|c c | c c }
Test trial & GPEKS & EP-GPEKS & GPADS & EP-GPADS\\
\hline
Trial 8 & 14.20 & 13.82 & 14.28 & 14.09\\
Trial 9 & 15.63 & 14.71 & 15.19 & 14.84\\
Trial 10 & 26.68 & 25.73 & 25.64 & 25.42
\end{tabular}
}
\vspace{-2mm}
\end{wrapfigure}
The computational demand the two inference methods for GPDSs we presented
is vastly different. High-dimensional approximate inference in the
motion capture example using moment matching (EP-GPADS) was about two
orders of magnitude slower than approximate inference based on
linearization of the posterior GP mean (EP-GPEKS): For updating the
posterior and the messages for a single time slice, the EP-GPEKS
required less than $\unit[0.5]{s}$, the EP-GPADS took about
$\unit[20]{s}$. Hence, numerical stability and more coherent posterior
inference with the EP-GPADS trade off against computational demands.

%
%



\section{Conclusion}
\label{sec:discussion}

We have presented an approximate message passing algorithm based on EP
for improved inference and Bayesian state estimation in GP dynamical
systems. Our message-passing formulation generalizes current inference
methods in GPDSs to iterative forward-backward smoothing.  This
generalization allows for improved predictions and comprises existing
methods for inference in the wider theory for dynamical systems as a
special case.  
Our new inference approach makes the full power of the GPDS model
available for the study of complex time-series data.  Future work
includes investigating alternatives to linearization and
moment matching when computing messages, and the more general problem
of learning in Gaussian process dynamical systems.


\section*{Acknowledgements}
The research leading to these results has received funding from the
European Community's Seventh Framework Programme (FP7/2007--2013) under
grant agreement \#270327 (CompLACS) and from the Canadian Institute
for Advanced Research (CIFAR). We thank Zhikun Wang for his help with
the motion capture data set.

\appendix
\label{sec:appendix}

\section{GP Predictions from Test Input 
\emph{Distributions}}  
\label{sec:appendix}
We will now review two approximations to the predictive distribution
\begin{align}
\prob(\vec x_{t})=\iint p(f(\vec x_{t-1})|\vec x_{t-1})\prob(\vec
x_{t-1})\d f\d\vec x_{t-1}\,,
\end{align}
where $f\sim\GP$ and $\vec
x_{t-1}\sim\gauss{\vec\mu_{t-1}}{\mat\Sigma_{t-1}}$.

\subsection{Moment Matching}
\label{sec:moment matching}
In the moment-matching approach, we analytically compute the mean
$\vec\mu_t$ and the covariance $\mat\Sigma_t$ of $\prob(\vec x_t)$.
Using the law of iterated expectations, we obtain
\begin{align}
\vec\mu_t = \E_{\vec x_{t-1}}\big[\E_f[f(\vec x_{t-1})
  |\vec x_{t-1}]\big] = \E_{\vec x_{t-1}}[m_f(\vec x_{t-1})]\,,
\end{align}
where $m_f$ is the posterior mean function of the dynamics GP. For
target dimension $a = 1,\dotsc,D$, we obtain
\begin{align}
  \vec\mu_t^a &= \vec q_a\T\vec\beta_a\,,\quad
q_{a_i} =
\tfrac{\sigma_f^2}{\sqrt{|\mat\Sigma_{t-1}\mat\Lambda_a\inv +
    \mat
    I|}}\exp\big(-\tfrac{1}{2}\vec\nu_i\T(\mat\Sigma_{t-1} +
\mat\Lambda_a)\inv\vec\nu_i\big) \,,\quad \vec\nu_i \coloneqq (\vec
x_i - \vec\mu_{t-1})
\label{eq:nu_i}
\end{align}
for $i = 1,\dotsc, n$, where $\vec\beta_a=\mat K_a\inv\vec y_a$.

Using the law of iterated variances, the entries of $\mat\Sigma_t$ for
target dimensions $a,b=1,\dotsc,D$ are
\begin{align}
  \hspace{-3mm}\sigma_{aa}^2 &\!=\! \E_{\vec x_{t-1}}\big[\var_f
  [\Delta_a|\vec x_{t-1}]\big]\!+\!
  \E_{f,\vec x_{t-1}}[\Delta_a^2]\!-\!(\vec\mu_t^a)^2,
  \label{eq:diagonal entry def.}\\
  \hspace{-3mm}\sigma_{ab}^2 &\!=\!
  \E_{f,\vec x_{t-1}}[\Delta_a\Delta_b]\!-\!\vec\mu_t^a\vec\mu_t^b\,,\quad
  a\neq b\,,
\label{eq:off-diagonal entry def}
\end{align}
respectively, where $\mu_t^a$ is known from \eq(\ref{eq:nu_i}). The
off-diagonal terms $\sigma_{ab}^2$ do not contain an additional term
$\E_{\vec x_{t-1}}[\cov_f [\Delta_a,\Delta_b|\vec x_{t-1}]]$ because
of the conditional independence assumption used for GP training:
Target dimensions do not covary for a \emph{given} $\vec x_{t-1}$.

For the term common to both $\sigma_{aa}^2$ and $\sigma_{ab}^2$, we
obtain
\begin{align}
  &\hspace{-1mm}\E_{f,\vec x_{t-1}}[\Delta_a\Delta_b]
  =\vec\beta_a\T\mat Q\vec\beta_b\,, \qquad Q_{ij}\!=\!\tfrac{k_a(\vec
    x_i,{\vec\mu}_{t-1})k_b(\vec x_j,
    {\vec\mu}_{t-1})}{\sqrt{|\mat R|}}\exp\big(\tfrac{1}{2}\vec
  z_{ij}\T\mat R\inv{\mat\Sigma}_{t-1}\vec z_{ij}\big)
\label{eq:off-diagonal intermediate result}
\end{align}
with $\mat R \coloneqq
{\mat\Sigma}_{t-1}(\mat\Lambda_a\inv+\mat\Lambda_b\inv)+\mat I$
and $\vec
z_{ij}\coloneqq\mat\Lambda_a\inv\vec\nu_i+\mat\Lambda_b\inv\vec\nu_j$
with $\vec \nu_i$ taken from \eq(\ref{eq:nu_i}).  Hence, the
\emph{off-diagonal} entries $\sigma_{ab}^2$ of $\mat\Sigma_t$ are
fully determined by \eqs(\ref{eq:nu_i}) and (\ref{eq:off-diagonal
  entry def}).

From \eq(\ref{eq:diagonal entry def.}), we see that the
\emph{diagonal} entries $\sigma_{aa}^2$ of $\mat\Sigma_t$ contain
an additional term
\begin{align}
  \hspace{-2mm}\E_{\vec x_{t-1}}\big[\var_f
  [\Delta_a|\vec x_{t-1}]\big]&= \sigma_{f_a}^2- \tr\big(\mat
  K_a\inv\mat Q\big) + \sigma_{w_a}^2
\label{eq:expected signal variance}
\end{align}
with $\mat Q$ given in \eq(\ref{eq:off-diagonal intermediate
  result}). This concludes the computation of $\mat\Sigma_t$.

The moment-matching approximation minimizes the KL divergence
KL$(p||q)$ between the true distribution $p$ and an approximate
Gaussian distribution $q$. This is generally a conservative
approximation, i.e., $q$ has probability mass where $p$ has
mass~\cite{Bishop2006}.

\subsection{Linearizing the GP Mean Function}
\label{sec:linearization}
An alternative way of approximating the predictive GP distribution for
\emph{uncertain} test inputs is to linearize the posterior GP mean
function~\cite{Ko2009}. This is equivalent to computing the expected
linearization of the GP distribution over functions.  Given this
linearized function, we apply standard results for mapping Gaussian
distributions through linear models.
Linearizing the posterior GP mean function yields to a predicted mean
that corresponds to the posterior GP mean function evaluated at the
mean of the input distribution, i.e.,
\begin{align}
  \vec\mu_t^a &= \E_f[f_a(\vec\mu_{t-1})] = \vec
  r_a\T\vec\beta_a\,,\qquad r_{a_i} =
  \sigma_{f_a}^2\exp\big(-\tfrac{1}{2} (\vec x_i -
  {\vec\mu}_{t-1})\T\mat\Lambda_a\inv (\vec x_i -
  {\vec\mu}_{t-1})\big)
\label{eq:r_a}
\end{align}
for $i=1,\dotsc,n$ and target dimensions $a = 1,\dotsc,D$, where $\vec
\beta_a=\mat K_a\inv\vec y_a$.
The covariance matrix $\mat\Sigma_t$ of the GP prediction is
\begin{align}
  \mat\Sigma_t &= \mat V{\mat\Sigma}_{t-1}\mat V\T +
  \mat\Sigma_w\,,\qquad
\mat V =
\tfrac{\partial\vec\mu_t}{\partial{\vec\mu}_{t-1}}=
\vec\beta_a\T\tfrac{\partial\vec
  r_a}{\partial{\vec\mu}_{t-1}}\,,
\label{eq:EKF pred covariance}
\end{align}
where $\vec r_a$ is given in \eq\eqref{eq:r_a} and $\mat V$ is the
Jacobian evaluated at $\vec\mu_{t-1}$. In \eq\eqref{eq:EKF pred
  covariance}, $\mat\Sigma_w$ is a diagonal matrix whose entries are
the model uncertainty plus the noise variance evaluated at
${\vec\mu}_{t-1}$. This means ``model uncertainty'' no longer depends
on the density of the data points. Instead it is assumed constant.

Using linearization, the approximation optimality in the KL sense of
the moment matching is lost. However, especially in high dimensions,
linearization is computationally more beneficial. This speedup is
largely due to the simplified treatment of model uncertainty.



\end{document}